\documentclass[10pt,twocolumn,letterpaper]{article}

\usepackage[pagenumbers]{cvpr} 

\usepackage{graphicx}
\usepackage{amsmath}
\usepackage{amssymb}
\usepackage{booktabs}
\usepackage{array}
\usepackage{amsmath}
\usepackage{enumitem}
\usepackage{algorithm}
\usepackage{algpseudocode}
\usepackage{bbm}
\usepackage[accsupp]{axessibility} 
\usepackage[pagebackref,breaklinks,colorlinks]{hyperref}

\usepackage{etoolbox} 

\usepackage[capitalize]{cleveref}
\crefname{section}{Sec.}{Secs.}
\Crefname{section}{Section}{Sections}
\Crefname{table}{Table}{Tables}
\crefname{table}{Tab.}{Tabs.}

\begin{document}

\title{Undoing the Damage of Label Shift for Cross-domain Semantic Segmentation}
\author{Yahao Liu \space\space\space\space Jinhong Deng \space\space\space\space Jiale Tao  \space\space\space\space Tong Chu\space\space\space\space  Lixin Duan \space\space\space\space  Wen Li\thanks{The corresponding author}   \\
        School of Computer Science and Engineering \& Shenzhen Institute for Advanced Study, \\University of Electronic Science and Technology of China
 \\
      {\tt\small \{lyhaolive, jhdeng1997, jialetao.std, uestcchutong, lxduan, liwenbnu \}@gmail.com}
   }
\maketitle

\begin{abstract}
Existing works typically treat cross-domain semantic segmentation~(CDSS) as a data distribution mismatch problem and focus on aligning the marginal distribution or conditional distribution. However, the label shift issue is unfortunately overlooked, which actually commonly exists in the CDSS task, and often causes a classifier bias in the learnt model. In this paper, we give an in-depth analysis and show that the damage of label shift can be overcome by aligning the data conditional distribution and correcting the posterior probability. To this end, we propose a novel approach to undo the damage of the label shift problem in CDSS. In implementation, we adopt class-level feature alignment for conditional distribution alignment, as well as two simple yet effective methods to rectify the classifier bias from source to target by remolding the classifier predictions. We conduct extensive experiments on the benchmark datasets of urban scenes, including GTA5 to Cityscapes and SYNTHIA to Cityscapes, where our proposed approach outperforms previous methods by a large margin. For instance, our model equipped with a self-training strategy reaches $59.3\%$ mIoU on GTA5 to Cityscapes, pushing to a new state-of-the-art. The code will be available at \href{https://github.com/manmanjun/Undoing_UDA}{https://github.com/manmanjun/Undoing\_UDA}.
\end{abstract}

\section{Introduction}
\label{sec:intro}
Semantic segmentation aims to classify every pixel in a given image. As a fundamental visual perception problem,  it is the basic module of many visual applications, such as autonomous driving systems. Remarkable progress in semantic segmentation has been made in recent years, driven by large-scale annotated datasets~\cite{cordts2016cityscapes,Caesar_2018_CVPR,8100027}. However, the massive and high-quality annotation, especially for semantic segmentation, can be costly and labor-intensive. Thus it is not always realistic to collect a sufficient number of well-annotated images for a new environment. Fortunately, we can obtain synthetic images with accurate pixel-level annotations rendered from the computer by a physical engine~\cite{richter2016playing,ros2016synthia}. However, the model trained with a synthetic domain often suffers from performance degradation since synthetic images~(source domain) and testing images~(target domain) are drawn from different distributions. This phenomenon commonly exists in cross-domain semantic segmentation~(CDSS) tasks, and many unsupervised domain adaptation models have been proposed to address this issue by transferring the knowledge from a label-rich source domain to the unlabeled target domain.

\begin{figure}[t]
\centering
   \includegraphics[width=1.0\linewidth]{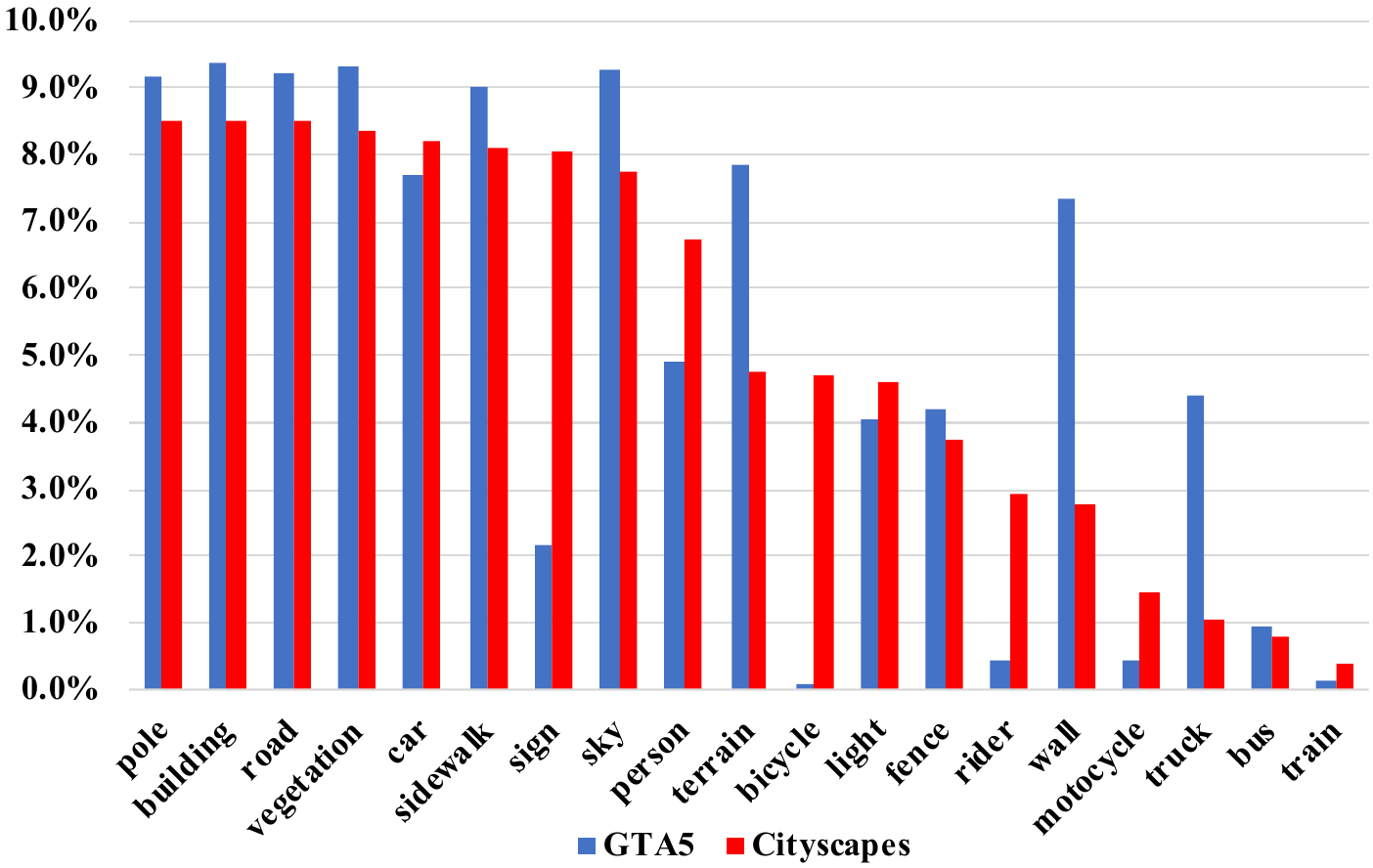}
   \caption{Label distribution in GTA5 and Cityscapes. There is an obvious label shift problem between the two datasets. For example, the frequency of ``rider'' in GTA5 is much less than that in Cityscapes while that of ``wall'' is opposite (Best viewed in color).}
\label{fig:label_stat}
\end{figure}

Most existing CDSS methods~\cite{Tsai_2018_CVPR,Chen_2018_CVPR,Luo_2019_CVPR, Paul_WeakSegDA_ECCV20,Du_2019_ICCV,wang_2020_ECCV} seek to learn domain-invariant representations via adversarial training to align marginal distributions~($p(x)$) or conditional distributions~($p(x|y)$). However, they ignore the label shift problem, which commonly exists in CDSS tasks, since the label distribution is often different across domains. 
As shown in Fig.~\ref{fig:label_stat}, taking the GTA5~\cite{richter2016playing} to Cityscapes~\cite{cordts2016cityscapes} as an example, the frequency of ``truck'' and ``wall'' in the source domain is higher than that in the target domain, and the frequency of ``bicycle'' is much lower than that in the target domain.

Therefore, in this work, we propose a novel approach to undo the damage of label shift in CDSS. First, we give an in-depth analysis and show that the classifier bias is the critical factor leading to the poor generalization ability of the learnt semantic segmentation model on the target domain when a label shift problem exists. It is inevitable that the classifier will be biased towards the source domain because the supervision signal comes only from it. At the same time, we show that the damage of label shift can be overcome by aligning the data conditional distribution and correcting the posterior probability. 

Then we adopt class-level feature alignment for conditional distribution alignment and propose two simple yet effective methods to rectify the classifier bias from source to target by remolding the classifier predictions.
In particular, after aligning the conditional distribution, we adjust the predictions of the classifier either in the training stage or the inference stage using the source and target label distribution. To achieve this, we design a new method to estimate the target label distribution using model predictions and the source pixel distribution prior. In this way, we can effectively address label shift, thus improving the model generalization ability on the target domain.

We conduct extensive experiments on the benchmark datasets of urban scenes, including GTA5 to Cityscapes and SYNTHIA to Cityscapes. By simply classifier rectification after aligning conditional distribution, our approach is remarkably effective. For example, for the GTA5 to Cityscapes task, it can reach $49.0\%$ mIoU, which surpasses all the adversarial based approaches. And our model can achieve $59.3\%$ mIoU after employing self-training. This validates the effectiveness of our proposed method.

In a nutshell, our contributions can be summarized as follows:
\begin{itemize}
\item 

We propose to address the label shift issue for CDSS tasks in a more realistic scenario~(\ie, the conditional distributions are different across domains) and reveal that the classifier bias is the critical factor leading to poor generalization on the target domain.

\item  We propose two simple yet effective methods to rectify the classifier bias from source to target by remolding the classifier predictions after explicitly aligning the conditional distribution.

\item We conduct extensive experiments on the benchmark datasets of urban scenes. The experimental results show that our approach outperforms other existing works, reaching a new state-of-the-art, validating our in-depth analysis on label shift.
\end{itemize}

\section{Related Work}
\subsection{Semantic Segmentation}
Semantic segmentation is a pixel-level prediction that can be used in a large number of visual applications, such as autonomous driving and security. Long~\cite{long2015fully} proposed the Fully Convolutional Network~(FCN), after which researchers started to focus on how to design the backbones for semantic segmentation, and proposed many excellent solutions such as UNet~\cite{ronneberger2015u}, SegNet~\cite{badrinarayanan2017segnet}, HRnet~\cite{wang2019deep}, and SegFormer~\cite{xie2021segformer}. Since the classifier plays an important role in improving the segmentation accuracy in semantic segmentation, many works delve into the mechanism of classifiers, including DeepLab~\cite{chen2017deeplab}, PSPnet~\cite{zhao2017pyramid} and OCRnet~\cite{YuanCW19}. However, although these approaches achieve impressive performance for semantic segmentation, they still require a large-scale well-annotated dataset, which is not easy to collect in real-world applications.

\subsection{Unsupervised Domain Adaptation}
Unsupervised Domain Adaptation~(UDA) methods aim to transfer the knowledge from a label-rich domain to an unlabelled domain and have been explored in many computer vision tasks, \eg, classification~\cite{long2017conditional,saito2018maximum}, object detection~\cite{chen2018domain,Deng_2021_CVPR}, and semantic segmentation~\cite{hoffman2016fcns,Tsai_2018_CVPR}. Most UDA methods are based on discrepancies between the source domain and target domain~(\textit{e.g.}, \emph{$\mathcal{H}$-divergence}~\cite{ben2010theory} or  \emph{$\mathcal{A}$-distance}~\cite{ben2007analysis}).  Among them, adversarial training is the most widely adopted strategy~\cite{ganin2016domain, long2017conditional, saito2018maximum, DMCD}. This methodology usually focuses on aligning the marginal distribution between the source domain and target domain under the assumption of covariate shift. Recently, researchers have found that this alignment strategy may lead to the phenomenon of misalignment~\cite{kumar2018co, chen2019progressive}. Therefore, some methods attempt to address the conditional shift between source and target domain by adversarial training~\cite{tachet2020domain, luo2021conditional}.

\subsection{Cross-domain Semantic Segmentation}

Exploiting synthetic data to help the real-world data has been a popular research topic in unsupervised domain adaptation. To address the cross-domain semantic segmentation problem, \cite{hoffman2016fcns, Tsai_2018_CVPR, Vu_2019_CVPR} use the domain adversarial learning with the structure information to align the marginal distribution. However, to eliminate the phenomenon of misalignment, many methods~\cite{Luo_2019_CVPR,Du_2019_ICCV,Paul_WeakSegDA_ECCV20,wang_2020_ECCV} have been proposed to conduct the conditional distribution alignment to learn class-level domain-invariant representation. 

Recently, self-supervised methods, including self-training~\cite{Zou_2018_ECCV,Zou_2019_ICCV,zheng2019unsupervised, zheng_2020_jour,mei_2020_ECCV,Zhang_2021_CVPR,What_Transferred_Dong_CVPR2020,9616392_Dong} and semantic consistency~\cite{Yang_2020_CVPR, tranheden2021dacs,Liu_2021_ICCV}, have been shown to significantly improve the cross-domain semantic segmentation performance.

However, the label shift problem between the source and target domain is still a long-standing but under-explored problem. Some approaches~\cite{10.5555/1642293.1642455,10.1016/j.neunet.2013.11.010} try to address this problem, but they typically assume that the conditional distribution is the same across domains, which is not realistic in practice. In a different approach than other studies, we propose to address the conditional shift and label shift simultaneously. Currently, the most related work is CLS~\cite{Liuxiao_2021_ICCV}, which proposes an algorithm to align the marginal distribution by adjusting the label distribution as an alternative to conditional distribution alignment. Different from it, we propose to conduct a two-stage learning paradigm by aligning conditional distribution and then considering label shift, which gives results in a better performance. Detailed analysis can be found in Section~\ref{sec:discussion}.
\section{Methodology}
In cross-domain semantic segmentation (CDSS), we are given a labeled source domain
$\mathcal{D}_{s}=\{(x_s^i, y_s^i)|_{i=1}^{N_s}\}$ where $x_s^i$ is an image and $y_s^i$ is the corresponding pixel-level annotation, as well as an unlabeled target domain $\mathcal{D}_{t}=\{x_t^i|_{i=1}^{N_t}\}$ where $x_t^i$ is a target image without annotation. Although the target domain is unlabeled, we generally assume the source and target domains share the same label space. For simplicity, we ignore the image size $H$ and $W$ in following loss function. The task of CDSS is to learn a semantic segmentation model $G$ which performs well on the target domain.

Generally, a segmentation network can be represented as $G = C \circ F$, where $F$ is the feature extractor and $C$ is the classifier. Existing methods~\cite{hoffman2016fcns,Tsai_2018_CVPR,Chen_2018_CVPR} mainly treated CDSS as a data distribution mismatch problem, and focused on training $F$ to align source and target domains by aligning either the marginal distribution~\cite{Tsai_2018_CVPR,Chen_2018_CVPR} or the conditional distribution~\cite{Luo_2019_CVPR, Paul_WeakSegDA_ECCV20,Du_2019_ICCV,wang_2020_ECCV}. 

However, data distribution mismatch is not the only problem of CDSS. There is also often a label distribution discrepancy between different domains. For example, the frequency of cars and buildings in the country is much lower than that in the urban scene, and the frequency of plants is higher in the country than it is in the city. More examples can be observed from Fig.~\ref{fig:label_stat}. Such a label distribution discrepancy will cause a classifier discrepancy between the source and target domains, and even though their data distributions are well aligned. In other words, we need to pay attention to the learnt classifier $C$ of the CDSS models to reduce such classifier bias caused by label distribution discrepancy. 

\begin{figure}
  \begin{center}
  \centering
    \includegraphics[width=1.0\linewidth]{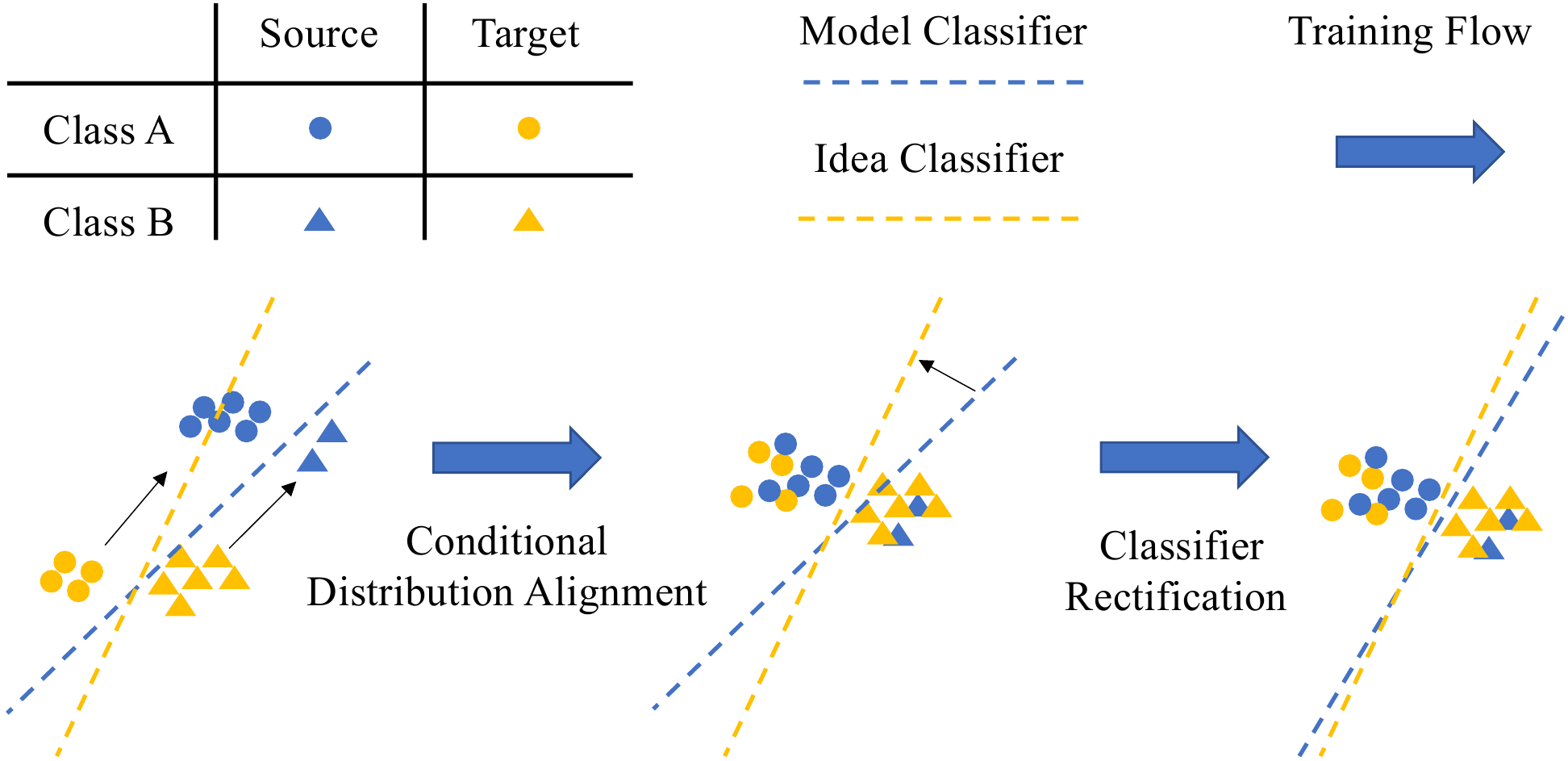}
    \caption{Overview of our proposed method. First, through conditional distribution alignment, the feature extractor will obtain conditional domain-invariant features. However, there is still a classifier discrepancy between the model classifier biased towards the source domain and the idea target classifier. Thus, we need to undo the damage of label shift for CDSS by classifier rectification.~(Best viewed in color).}
  \label{fig:diagram}
  \vspace{-2mm}
\end{center}
\end{figure}

To this end, we propose a novel approach to simultaneously handle the data distribution discrepancy and the label distribution discrepancy shown in Fig.~\ref{fig:diagram}, in which we exploit the conditional distribution alignment strategy and design two label distribution adjustment strategies for this purpose. In the following, we first present a rigorous analysis on the classifier bias and label distribution discrepancy in Section ~\ref{sec:mot}, then follow with the conditional distribution alignment strategy in Section~\ref{sec:cda} and label distribution adjustment strategies in Section~\ref{sec:clsa}. Additional details and discussions are presented in Section~\ref{sec:lde} and \ref{sec:discussion}.

\subsection{Motivation}
\label{sec:mot}

Given a segmentation model $G = C \circ F$, the classifier $C$ is expected to output the posterior probability in the ideal case, \ie, $G(x) = C(F(x)) = p(Y|F(x))$ where $x$ is an image, and $Y$ is a random variable representing the semantic label.
From Bayes' theorem, we have:
 \begin{equation}
C (F(x))=P(Y|F(x))=\frac{P(F(x)|Y) P(Y)}{P(F(x))},
\end{equation}
where $p(F(x))$ is a constant, and $x$ is a sampled image. We can observe that the classifier is affected by the conditional distribution $P(F(x)|Y)$ and the label distribution $P(Y)$.

Correspondingly, for a segmentation model $G_{s} = C_{s} \circ F_s$ trained on the source data, the output of $C_{s}$ meets the following relationship:
\begin{equation}
C_{s}(F(x)) \propto P_{s}(F_s(x)|Y) P_{s}(Y).
\label{eq:src_classifier}
\end{equation}

Similarly, for the target domain, an ideal segmentation model $G_{t} = C_{t} \circ F_t$  shall satisfy,
\begin{equation}
C_{t}\left(F_t\left(x\right)\right) \propto P_{t}\left(F_t\left(x\right)|Y\right) P_{t}\left(Y\right).
\label{eq:tgt_classifier}
\end{equation}

In CDSS, our goal is to adapt a model trained with source supervision to the target domain. Without loss of generality, we assume the feature extractor is identical for two domains, and unify them as $F = F_s =F_t$. Then, we further assume that the conditional data distribution $P(X|Y)$ is well aligned, \ie, we have $ P_{s}(F(x) \mid Y)=P_{t} (F(x) \mid Y)$. Using Eq.~(\ref{eq:src_classifier}) and Eq.~(\ref{eq:tgt_classifier}), the relation of $C_{s}$ and $C_{t}$ can be derived as:
\begin{equation}
C_{t}(F(x)) \propto \frac{C_{s}(F(x)) P_{t}(Y)}{P_{s}(Y)}.
\label{eq:tgt_src_classifier_relation}
\end{equation}

We can observe from Eq.~(\ref{eq:tgt_src_classifier_relation}) that the label distribution discrepancy between the source and target domains would cause a classifier bias issue. It also points out a way to rectify the source classifier to approach the ideal target classifier. We detail the solution in the following subsections. 


\subsection{Conditional Distribution Alignment}
\label{sec:cda}

As shown above, a presumption for rectifying the classifier bias using Eq.~(\ref{eq:tgt_src_classifier_relation}) is the data conditional distribution is well aligned. Next, we discuss how to perform conditional distribution alignment in CDSS. 

Adversarial training is a common strategy to align two distributions in CDSS, where a domain discriminator $D$ is applied to learn domain-invariant features. The discriminator $D$ adapts the features and tries to identify the domain label for the inputted images while the feature extractor $F$ attempts to extract domain-invariant features. This is achieved by a Gradient Reversal Layer~(GRL) between the feature extractor $F$ and the discriminator $D$ or by employing an alternative optimization in a min-max manner. 

For conditional distributional alignment, a few solutions~\cite{Luo_2019_CVPR, Paul_WeakSegDA_ECCV20,Du_2019_ICCV,wang_2020_ECCV} were also proposed. Generally, they are implemented by integrating the label information into the domain discriminator $D$. For example, FADA~\cite{wang_2020_ECCV} proposed to extend the domain discriminator to output both the domain label and the class label. Specifically, they use a discriminator with $2K$-dimensional output, representing the class conditional domain probability. Formally, the optimization process of the adversarial learning process can be written as:
\begin{equation}
\label{eq:l_seg_l_adv}
\min _{F,C} \mathcal{L}_{seg} + \lambda_{adv}\mathcal{L}_{adv} ,
\end{equation}
\begin{equation}
\label{eq:l_d}
\min _{D} \mathcal{L}_{D} ,
\end{equation}
where $\mathcal{L}_{seg}$ is the cross-entropy loss for source domain, $\mathcal{L}_{D}$ is designed to train the discriminator $D$, $\mathcal{L}_{a d v}$ is used to make $F$ extract the conditional domain invariant feature. They are defined as follows:
\begin{equation}
    \mathcal{L}_{seg}=-\sum_{i=1}^{N_{s}} \sum_{k=1}^{K} y^{i k}_{s} \log \left(C\left(F\left(x_s^{i}\right)\right)\right),
\end{equation}

\begin{equation}
\mathcal{L}_{a d v}=-\sum_{j=1}^{N_{t}} \sum_{k=1}^{K} a^{j k}_{t} \log D\left(d=0,  y=k \mid F(x^{j}_{t})\right),\label{eq:fada_adv}
\end{equation}

\begin{equation}
\begin{aligned}
\mathcal{L}_{D}=&-\sum_{i=1}^{N_{s}} \sum_{k=1}^{K} a^{i k}_{s} \log D (d=0,  y=k \mid F(x^{i}_{s}) ) \\
&-\sum_{j=1}^{N_{t}} \sum_{k=1}^{K} a^{j k}_{t} \log D (d=1,  y=k \mid F(x^{j}_{t})),
\label{eq:fada_d}
\end{aligned}
\end{equation}
where $a^{i k}_{s}$ and  $a^{j k}_{t}$ are the $k$-th class knowledge for the different domains and $d$ is the domain label, $0$ represents the source domain and $1$ represents the target domain. Please refer to~\cite{wang_2020_ECCV} for more details. In this way, the feature extractor will obtain conditional domain-invariant features after training the model using the conditional distribution alignment strategy. We simply employ FADA for conditional data distribution alignment and focus on validating the importance of classifier bias correction.

\subsection{Classifier Rectification}
\label{sec:clsa}
After aligning the conditional distribution $P(X|Y)$, we are able to use Eq.~(\ref{eq:tgt_src_classifier_relation}) to rectify the source classifier to be the target one, \ie,  through adjusting the predictions of $C_s$ by the ratio $P_{t}(Y) / P_{s}(Y)$. We propose
two strategies, classifier refinement and inference adjustment, for this purpose. 

\paragraph{Classifier Refinement~(CR)} 
We consider how to train a target classifier using the source supervision in CDSS. In particular, supposing we have a classifier $C_t$, according to Eq.~(\ref{eq:tgt_src_classifier_relation}), we have,
\begin{equation}
C_{s}\left(F(x)\right) \propto \frac{C_{t}\left(F\left(x\right)\right) P_{s}\left(Y\right)}{P_{t}\left(Y\right)}.
\label{eq:tgt_src_classifier_relation2}
\end{equation}

This implies that when training the segmentation model with source supervision, if we remold the predictions of the classifier using the above formulation, we are able to obtain a target classifier that can be directly used for predicting target domain samples. 

Specifically, after performing the conditional distribution alignment, we first fix the backbone such that the conditional distribution of features remains unchanged. Then we continually train the classifier using labeled source sample by minimizing the semantic segmentation loss as follows:
\begin{equation}
\min _{C_{cda}} \mathcal{L}^{'}_{\text {seg}},
\label{eq:refining the classifier_1}
\end{equation}
\begin{equation}
    \mathcal{L}^{'}_{\text {seg}}=-\sum_{i=1}^{n_{s}} \sum_{k=1}^{K} y^{i k}_{s} \log (\hat{p}^{i k}_{s}),
\label{eq:refining the classifier_2}
\end{equation}
\begin{equation}
\begin{split}
\hat{p}^{i k}_{s}=  \frac{p^{i k}_{s} \cdot \frac{P_{s}(Y=k)}{P_{t}(Y=k)}}{ \sum_{k^{'}=1}^{K} \left(p^{i k^{'}}_{s} \cdot \frac{P_{s}(Y=k^{'})}{P_{t}(Y=k^{'})}\right)} .
\label{eq:refining the classifier_3}
\end{split}
\end{equation}

 \paragraph{Inference Adjustment~(IA):} An alternative idea is that we directly adjust the output of the existing network $G_{cda}$ in the inference stage on the target domain. Since the conditional probability distribution is now aligned, the hypothesis of label shift is satisfied. So according to the Eq.~(\ref{eq:tgt_src_classifier_relation}), we adjust the output $p^{i}_{t}$ of the existing network $G_{cda}$ in the inference stage of the target domain data $x^{i}$ to obtain the final prediction $y_{IA}$. Formally, we adjust the predictions of the classifier as follows:
\begin{equation}
\begin{split}
y_{IA}^{i}=\operatorname{argmax} _{k}\left(p^{i}_{t} \cdot \frac{P_{t}(Y)}{P_{s}(Y)}\right) .
\end{split}
\end{equation}

In CDSS, both CR and IA are from the perspective of the classifier to correct the problem caused by domain shift. CR makes the classifier biased towards the target domain at the training stage, while IA aims to directly change the output of the model for the target domain at the inference stage. Both strategies are essentially the same in theory, and one can choose either way in combination with the data distribution alignment. 

Moreover, it is popular in existing state-of-the-art CDSS methods~\cite{zheng_2020_jour,mei_2020_ECCV,Zhang_2021_CVPR} to use the domain alignment model as a warm-up model for generating pseudo-labels and then perform self-training. As our proposed strategies are able to improve the performance of adversarial based CDSS models, the quality of pseudo-labels in these self-training methods will also be improved, thus further boosting the current state-of-the-art CDSS methods.
\subsection{Label Distribution Estimation}
\label{sec:lde}
Recall that when correcting the classifier using the two strategies in Section~\ref{sec:clsa}, the prior label distribution $P_s(Y)$  and $P_t(Y)$ are required. We discuss how to estimate them below. 

For the source domain, as the labels are available,  we directly obtain the label distribution estimation $P_s(\hat{Y})$ by counting the image-level class labels from the ground truth label throughout the whole dataset. Specifically, we denote the count of the image-level label of $i$-th source image as $I_{s}^i(k)$. We can obtain $I_{s}^i(k)$ as follows:
\begin{eqnarray}
I_{s}^i(k)= \mathbbm{1}\left[\left(\sum_{h=1}^{H} \sum_{w=1}^{W}\mathbbm{1}\left[y_{s}^{i}\left(h, w\right)==k\right]\right)>n_{s}\right],
\end{eqnarray}
where $n_s$ means that we ignore the labels that only exist very few pixels in an image and $\mathbbm{1}[\cdot]$ is an indicator function that equals $1$ when $[\cdot]$ is true, otherwise 0. The image-level source label distribution can be estimated for the whole dataset as follows:
\begin{equation}
P_{s}(\hat{Y}=k)=\frac{\sum_{i=1}^{N_{s}}I_{s}^i(k)}{\sum_{k=1}^{K}\sum_{i=1}^{N_{s}}I_{s}^i(k)}.
\label{eq:source_label_estimation}
\end{equation}

For the target domain without ground truth, we use the conditional alignment model $G_{cda}$ to estimate the image-level category probabilities $p_{i}(Y)$ by the smooth max pooling strategy~\cite{pinheiro2015image, Paul_WeakSegDA_ECCV20, Lv_2020_CVPR}. Then, we compare it with the pixel ratio $p_{pix}(Y)$ of each category in the source domain dataset. For the class $k$, if $p_{i}(k)$ is greater than $p_{pix}(k)$, it is deemed that the category appears in the image, otherwise $I_{t}^i(k)$ is $0$. After dataset normalization, we can obtain the label distribution estimation $P_{t}(\hat{Y})$ of the target domain. The details are as follows:
\begin{equation}
p_{i}(k)=\log \left[\frac{1}{H W}\sum\limits_{h=1}^{H} \sum\limits_{w=1}^{W} \exp\left(G_{cda}\left(X^i_t\right)\left(h, w, k\right)\right)\right],
\end{equation}
\begin{equation}
I_{t}^i(k)=\mathbbm{1}\left[p_{i}(k) > p_{pix}(k)\right],
\end{equation}
\begin{equation}
P_{t}(\hat{Y}=k)=\frac{\sum_{i=1}^{N_{t}}I_{t}^i(k)}{\sum_{k=1}^{K}\sum_{i=1}^{N_{t}}I_{t}^i(k)}.
\label{eq:target_label_estimation}
\end{equation}

\subsection{Discussion}
\label{sec:discussion}
In a recent study, CLS~\cite{Liuxiao_2021_ICCV} was proposed to address the label shift issue in the unsupervised domain adaptation task. While they also validate their approach for CDSS in experiments, the performance is less satisfactory compared to that of ours (see details in Section~\ref{sec:exp}). We analyze the potential reasons as follows. 

In CLS, they showed that conditional distribution alignment can be achieved by merging the class classifier and the domain classifier together and re-weight the class classifier loss with the label distribution ratio. After that, a posterior alignment is performed to correct the trained class classifier.

We clarify the difference between CLS and ours to gain a better understanding of the label shift problem for CDSS. In particular, given a source image $x_s^i$ with label $k$, the modified classifier loss can be written as:
\begin{equation}
\mathcal{L}(k,x_s^i) = \frac{P_{t}(Y=k)}{P_{s}(Y=k)}log\left[1+\sum_{k^{\prime}\not=k}\frac{p^{ik^{\prime}}_{s}}{p^{ik}_{s}}\right].
\label{eqn:cls_loss}
\end{equation}

For convenience, we rewrite the loss of classifier refinement in Eq.~(\ref{eq:refining the classifier_2}) as follows:
\begin{equation}
\begin{split}
L(k,x_s^i) = log\left[1+ \sum_{k^{\prime}\not=k}\frac{P_{s}(Y=k^{\prime})}{P_{s}(Y=k)}\frac{P_{t}(Y=k)}{P_{t}(Y=k^{\prime})}\frac{p^{ik^{\prime}}_{s}}{p^{ik}_{s}}\right].
\end{split}
\end{equation}

It can be observed that the difference is that CLS applies the label distribution ratio as the loss weight while we directly adjust the posterior probability output from the classifier. As stated in ~\cite{menon2021longtail,byrd2019effect}, while re-weigting the loss with the label distribution ratio also helps to guide the classifier to approach the target classifier, yet not as effectively as directly using it to correct the classifier output like our classifier refinement loss. 

Another important difference is that CLS utilizes the label distribution ratio in both the training and inference phases. As discussed in Section~\ref{sec:clsa}, the label distribution ratio is able to recover the target classifier in either phase if it is correctly applied. Using it in both phases is not desirable from the probabilistic perspective. We conjecture that due to the loss weighting in the training phase is less effective, they might hope to compensate for it by applying the label distribution ratio again at the inference phase. 

On the contrary, in our approach, we clearly disentangle the conditional distribution alignment and label shift correction issues. By applying the label distribution ratio in either the classifier refinement or the inference adjustment, we achieve satisfactory improvement on segmentation performance.


\begin{table*}[t]
\caption{Results~(in $\%$) of adapting GTA5 to Cityscapes. All the results are obtained from the ResNet-101-based models. Except ProDA~\cite{Zhang_2021_CVPR} and R-MRNet~\cite{zheng_2020_jour} using the modified ASPP as the classifier, other methods use the original ASPP.}
\label{Tab:GTA5}
\centering
\resizebox{\textwidth}{!}{
    \setlength{\tabcolsep}{1pt}
    \begin{tabular}{p{2.6cm}|p{0.9cm}<{\centering}p{0.9cm}<{\centering}p{0.9cm}<{\centering}p{0.9cm}<{\centering}p{0.9cm}<{\centering}p{0.9cm}<{\centering}p{0.9cm}<{\centering}p{0.9cm}<{\centering}p{0.9cm}<{\centering}p{0.9cm}<{\centering}p{0.9cm}<{\centering}p{0.9cm}<{\centering}p{0.9cm}<{\centering}p{0.9cm}<{\centering}p{0.9cm}<{\centering}p{0.9cm}<{\centering}p{0.9cm}<{\centering}p{0.9cm}<{\centering}p{0.9cm}<{\centering}|p{0.9cm}<{\centering}} \toprule
    Method& \rotatebox{90}{road}& \rotatebox{90}{ sidewalk}& \rotatebox{90}{ building}& \rotatebox{90}{ wall}& \rotatebox{90}{ fence}& \rotatebox{90}{ pole}& \rotatebox{90}{ light}& \rotatebox{90}{ sign}& \rotatebox{90}{ vege.}& \rotatebox{90}{ terrain}& \rotatebox{90}{ sky}& \rotatebox{90}{ person}& \rotatebox{90}{ rider}& \rotatebox{90}{ car}& \rotatebox{90}{ truck}& \rotatebox{90}{ bus}& \rotatebox{90}{ train}& \rotatebox{90}{ motor}& \rotatebox{90}{ bike}& mIoU\\
    \hline \hline
    Source&  65.0&16.1&68.7&18.6&16.8&21.3&31.4&11.2&83.0&22.0&78.0&54.4&33.8&73.9&12.7&30.7&13.7&28.1&19.7&36.8\\
    AdaptSegNet~\cite{Tsai_2018_CVPR}&86.5&36.0&79.9&23.4&23.3&23.9&35.2&14.8&83.4&33.3&75.6&58.5&27.6&73.7&32.5&35.4&3.9&30.1&28.1&42.4\\
    AdvEnt~\cite{Vu_2019_CVPR}& 89.4& 33.1& 81.0& 26.6& 26.8& 27.2& 33.5& 24.7& 83.9& 36.7& 78.8& 58.7& 30.5& 84.8& 38.5& 44.5& 1.7& 31.6& 32.4& 45.5\\
    CLAN~\cite{Luo_2019_CVPR}&87.0&27.1&79.6&27.3&23.3&28.3&35.5&24.2&83.6&27.4&74.2&58.6&28.0&76.2&33.1&36.7&6.7&31.9&31.4&43.2\\

    FADA~\cite{wang_2020_ECCV}&87.0&37.6&83.3&36.9&25.3&30.9&35.3&21.0&82.7&36.8&83.1&58.3&34.1&83.3&31.5&35.0&24.4&34.3&32.0&46.9 \\ \hline
     Our IA &87.9&37.0&83.3&37.0&25.0&31.0&35.7&24.9&83.4&38.9&85.7&58.0&35.4&83.6&35.3&36.3&30.7&32.5&45.2&48.8\\  
    Our CR &89.1&34.3&83.6&38.3&27.5&28.9&34.7&17.6&84.2&41.0&85.1&57.8&33.7&85.1&38.5&41.3&30.7&31.1&48.0&49.0\\ \hline  
    FDA~\cite{Yang_2020_CVPR}& 92.5& 53.3& 82.4& 26.5& 27.6& 36.4& 40.6& 38.9& 82.3& 39.8& 78.0& 62.6& 34.4& 84.9& 34.1& 53.1& 16.9& 27.7& 46.4& 50.5\\
    DACS~\cite{tranheden2021dacs}& 89.9& 39.7& \textbf{87.9}& 30.7& 39.5& 38.5& 46.4& 52.8& 88.0& 44.0& 88.8& 67.2& 35.8& 84.5& 45.7& 50.2& 0.0& 27.3& 34.0& 52.1\\
    CRST~\cite{Zou_2019_ICCV}& 91.0& 55.4& 80.0& 33.7& 21.4& 37.3& 32.9& 24.5& 85.0& 34.1& 80.8& 57.7& 24.6& 84.1& 27.8& 30.1& 26.9& 26.0& 42.3& 47.1\\
    FADA+SD~\cite{wang_2020_ECCV}&92.5&47.5&85.1&37.6&32.8&33.4&33.8&18.4&85.3&37.7&83.5&63.2&39.7&87.5&32.9&47.8&1.6&34.9&39.5&49.2\\
    IAST~\cite{mei_2020_ECCV}& 93.8& 57.8& 85.1& 39.5& 26.7& 26.2& 43.1& 34.7& 84.9& 32.9& 88.0& 62.6& 29.0& 87.3& 39.2& 49.6& 23.2& 34.7& 39.6& 51.5\\ 
    CLS~\cite{Liuxiao_2021_ICCV}+IAST&\textbf{94.7}&60.1&85.6&39.5&24.4&44.1&39.5&20.6&88.7&38.7&80.3&67.2&35.1&86.5&37.0&45.4&\textbf{39.0}&37.9&46.2&53.0 \\
    \hline
    Ours+SD &91.2&45.1&85.5&41.0&30.8&36.0&41.1&19.3&87.4&45.7&88.7&64.4&37.8&87.5&41.8&51.2&11.2&41.6&54.9&52.7\\ 
    Ours+IAST &94.1&\textbf{61.3}&86.5&39.3&33.5&38.3&48.9&38.5&87.2&44.2&89.3&63.4&38.3&86.2&30.5&43.0&33.6&43.1&54.8&55.5\\ \hline    R-MRNet~\cite{zheng_2020_jour}& 90.4& 31.2& 85.1& 36.9& 25.6& 37.5& 48.8 & 48.5& 85.3& 34.8& 81.1& 64.4& 36.8& 86.3& 34.9& 52.2& 1.7& 29.0& 44.6& 50.3\\
    ProDA~\cite{Zhang_2021_CVPR}&87.8&56.0&79.7&\textbf{46.3}&\textbf{44.8}&\textbf{45.6}&53.5&\textbf{53.5}&88.6&45.2&82.1&70.7&\textbf{39.2}&88.8&45.5&59.4&1.0&\textbf{48.9}&56.4&57.5 \\
    \hline
    Ours+ProDA & 92.9& 52.7 & 87.2& 39.4& 41.3& 43.9& \textbf{55.0}& 52.9& \textbf{89.3}& \textbf{48.2}& \textbf{91.2}& \textbf{71.4}& 36.0& \textbf{90.2}& \textbf{67.9} & \textbf{59.8} & 0.0& 48.5& \textbf{59.3}& \textbf{59.3}\\
    \hline
    \end{tabular}
}
\end{table*}


\section{Experiments}
\label{sec:exp}
\subsection{Datasets}
We conduct experiments on two common CDSS benchmarks: GTA5 to Cityscapes and SYNTHIA to Cityscapes. They are both synthetic-to-real scenarios. GTA5 and SYNTHIA are synthetic datasets used as source domains and Cityscapes is a real dataset as the target domain.
\begin{itemize}[itemsep=0pt]
    \item \textbf{Cityscapes}~\cite{cordts2016cityscapes} is a real-world urban scene semantic segmentation benchmark dataset. Following~\cite{hoffman2016fcns, Tsai_2018_CVPR, Vu_2019_CVPR}, we use the 2,975 images from its training set without the annotation as the unlabeled target samples and evaluate our method on its validation set which includes 500 images.
  \item \textbf{GTA5}~\cite{richter2016playing} is a synthetic dataset automatically rendered from a computer game named Grand Theft Auto V~(GTA5). We use all of its 24,966 images as the source domain, which consists of 19 classes in common with Cityscapes.
    \item \textbf{SYNTHIA}~\cite{ros2016synthia} is also a popular synthetic semantic segmentation benchmark dataset. We use its subset SYNTHIA-RAND-CITYSCAPES as the source domain, which contains $9,400$ images and 16 classes in common with Cityscapes.
\end{itemize}

\subsection{Implementation Details}
In our experiments, following the methods of previous studies~\cite{Tsai_2018_CVPR, Vu_2019_CVPR,Luo_2019_CVPR,Du_2019_ICCV,wang_2020_ECCV,Zou_2018_ECCV,Zou_2019_ICCV,zheng2019unsupervised, zheng_2020_jour,mei_2020_ECCV,Zhang_2021_CVPR,Yang_2020_CVPR, tranheden2021dacs, guo2021metacorrection}, we take the ResNet-101~\cite{he2016deep} model pretrained on ImageNet~\cite{deng2009imagenet} as the backbone. For a fair comparison, we conduct experiments with the original Atrous Spatial Pyramid Pooling~(ASPP) in Deeplab-v2~\cite{chen2017rethinking} and the modified ASPP~\cite{zheng2019unsupervised,zheng_2020_jour} as the segmentation classifier. When performing the conditional adversarial learning, we choose FADA~\cite{wang_2020_ECCV} as our baseline. For the original ASPP, we use checkpoints provided by FADA. For methods that use the modified ASPP, we reproduce experiments for FADA as a new baseline. In our experiments, we use the Stochastic Gradient Descent~(SGD) optimizer with the momentum is set to $0.9$ and the weight decay is set to $10^{-4}$. We employ the polynomial decay with power of $0.9$ with the initial learning rate of $2.5\times10^{-4}$. At the self-training stage, we follow the corresponding training strategy of the pseudo label methods~\cite{mei_2020_ECCV,Zhang_2021_CVPR}. We use Intersection over Union~(IoU) as the evaluation metric and report per-class IoU and mean IoU over all classes. We conduct all the experiments on Tesla V100 GPUs with a PyTorch implementation.
\begin{table*}[t]
\caption{Results~(in $\%$) of adapting SYNTHIA to Cityscapes. \text{mIoU*} denotes the mean IoU over 13 classes excluding those marked with *. Classes not evaluated are replaced by '-'. All the results are generated from the ResNet-101-based models. Except ProDA~\cite{Zhang_2021_CVPR} and R-MRNet~\cite{zheng_2020_jour} using the modified ASPP as the classifier, other methods use the original ASPP.}
\label{Tab:SYNTHIA}
\centering
\resizebox{\textwidth}{!}{
    \setlength{\tabcolsep}{1pt}
    \begin{tabular}{p{2.6cm}|p{0.9cm}<{\centering}p{0.9cm}<{\centering}p{0.9cm}<{\centering}p{0.9cm}<{\centering}p{0.9cm}<{\centering}p{0.9cm}<{\centering}p{0.9cm}<{\centering}p{0.9cm}<{\centering}p{0.9cm}<{\centering}p{0.9cm}<{\centering}p{0.9cm}<{\centering}p{0.9cm}<{\centering}p{0.9cm}<{\centering}p{0.9cm}<{\centering}p{0.9cm}<{\centering}p{0.9cm}|p{1cm}<{\centering}p{0.9cm}<{\centering}} \toprule
    Method& \rotatebox{90}{road}& \rotatebox{90}{ sidewalk}& \rotatebox{90}{ building}& \text{\rotatebox{90}{ wall*}}& \text{\rotatebox{90}{ fence*}}& \text{\rotatebox{90}{ pole*}}& \rotatebox{90}{ light}& \rotatebox{90}{ sign}& \rotatebox{90}{ vege.}& \rotatebox{90}{ sky}& \rotatebox{90}{ person}& \rotatebox{90}{ rider}& \rotatebox{90}{ car}&  \rotatebox{90}{ bus}&  \rotatebox{90}{ motor}& \rotatebox{90}{ bike}& \text{mIoU*}& mIoU\\
    \hline \hline
    Source& 55.6&23.8&74.6&9.2&0.2&24.4&6.1&12.1&74.8&79.0&55.3&19.1&39.6&23.3&13.7&25.0&38.6&33.5 \\
    AdaptSegNet~\cite{Tsai_2018_CVPR}&81.7 &39.1 &78.4 &11.1&0.3&25.8&6.8&9.0&79.1&80.8&54.8&21.0&66.8&34.7&13.8&29.9&45.8&39.6 \\
    AdvEnt~\cite{Vu_2019_CVPR}& 85.6& 42.2& 79.7& 8.7& 0.4& 25.9& 5.4& 8.1& 80.4& 84.1& 57.9& 23.8& 73.3& 36.4& 14.2& 33.0& 48.0& 41.2\\
    CLAN~\cite{Luo_2019_CVPR}& 81.3& 37.0& 80.1&  -&  -&  -& 16.1& 13.7& 78.2& 81.5& 53.4& 21.2& 73.0& 32.9& 22.6& 30.7& 47.8& - \\
    FADA~\cite{wang_2020_ECCV}&81.3&35.1&80.8&9.6&0.2&26.8&9.1&17.8&82.4&81.5&49.9&18.8&78.9&33.3&15.3&33.7& 47.5 &40.9 \\ \hline
    Our IA &82.2&35.6&80.8&9.0&0.2&27.1&12.4&21.3&82.3&80.7&54.4&21.2&80.0&36.6&14.0&42.2&49.5&42.5 \\
    Our CR &83.6&36.2&80.9&10.3&0.1&27.4&17.6&22.8&81.5&81.2&54.6&20.1&80.3&38.1&11.1&42.9&50.1&43.0 \\ \hline 
    FDA~\cite{Yang_2020_CVPR}& 79.3& 35.0& 73.2&  -&  -&  -& 19.9& 24.0& 61.7& 82.6& 61.4& \textbf{31.1}& 83.9& 40.8& 38.4& 51.1& 52.5&  -\\
    DACS~\cite{tranheden2021dacs}& 80.6& 25.1& 81.9& 21.5& 2.9& 37.2& 22.7& 24.0& 83.7& \textbf{90.8}& 67.6& 38.3& 82.9& 38.9& 28.5& 47.6& 54.8& 48.3\\

    CRST~\cite{Zou_2019_ICCV}& 67.7& 32.2& 73.9& 10.7& 1.6& 37.4& 22.2& 31.2& 80.8& 80.5& 60.8& 29.1& 82.8& 25.0& 19.4& 45.3& 50.1& 43.8\\
    FADA+SD~\cite{wang_2020_ECCV}& 84.5& 40.1& 83.1& 4.8& 0.0& 34.3& 20.1& 27.2& 84.8& 84.0& 53.5& 22.6& 85.4& 43.7& 26.8& 27.8& 52.5& 45.2\\    
    IAST~\cite{mei_2020_ECCV}& 81.9& 41.5& 83.3& 17.7& \textbf{4.6}& 32.3& 30.9& 28.8& 83.4& 85.0& 65.5& 30.8& 86.5& 38.2& 33.1& 52.7& 57.0& 49.8\\    \hline 

    Ours+SD &86.9&42.9&83.3&9.9&0.0&35.3&17.2&26.0&85.4&83.0&62.0&18.5&86.7&51.4&12.8&50.0&54.3&47.0 \\

    Ours+IAST &84.6&43.0&84.1&\textbf{38.1}&0.5&36.7&32.9&36.2&83.1&81.9&65.6&33.4&80.5&34.5&38.2&53.1&57.8&51.6 \\ \hline
    R-MRNet~\cite{zheng_2020_jour}& 87.6& 41.9& 83.1& 14.7& 1.7& 36.2& 31.3& 19.9& 81.6& 80.6& 63.0& 21.8& 86.2& 40.7& 23.6& 53.1& 54.9& 47.9\\
    ProDA~\cite{Zhang_2021_CVPR}&\textbf{87.8}&\textbf{45.7}&\textbf{84.6}&37.1&0.6&44.0&54.6&37.0&\textbf{88.1}&84.4&\textbf{74.2}&24.3&88.2&51.1&40.5&45.6&62.0&55.5 \\
    \hline 
    Ours+ProDA & 82.5&37.2&81.1&23.8&0.0&\textbf{45.7}&\textbf{57.2}&\textbf{47.6}&87.7&85.8&74.1&28.6&\textbf{88.4}&\textbf{66.0}&\textbf{47.0}&\textbf{55.3}&\textbf{64.5}&\textbf{56.7} \\
    \hline
\end{tabular}
}
\end{table*}

\subsection{Analysis}
\paragraph{Comparisons with state-of-the-art methods} 
 We compare our proposed methods with previous state-of-the-art CDSS approaches on the GTA5 to Cityscapes and SYNTHIA to Cityscapes in Table~\ref{Tab:GTA5} and Table~\ref{Tab:SYNTHIA}, respectively. The CDSS methods can be divided into three categories: 1) domain alignment methods including AdaptSegNet~\cite{Tsai_2018_CVPR}, AdvEnt~\cite{Vu_2019_CVPR}, CLAN~\cite{Luo_2019_CVPR}, and FADA~\cite{wang_2020_ECCV}, 2) self-training methods including CRST~\cite{Zou_2019_ICCV}, R-MRNet~\cite{zheng_2020_jour}, IAST~\cite{mei_2020_ECCV}, and ProDA~\cite{Zhang_2021_CVPR},
3) the data argument methods including FDA~\cite{Yang_2020_CVPR}, DACS~\cite{tranheden2021dacs}. All the models are trained using ResNet-101 as the backbone and original ASPP as the classifier except that ProDA~\cite{Zhang_2021_CVPR} and R-MRNet~\cite{zheng_2020_jour}, which use the modified ASPP as the classifier. Thus, we reproduce FADA~\cite{wang_2020_ECCV} based on the modified ASPP when compared with ProDA and R-MRNet. It is noted that our work primarily addresses the label shift after the conditional distribution alignment. Thus, it is clear that our method obtains comparable or greater results compared with domain alignment methods~\cite{Tsai_2018_CVPR,Vu_2019_CVPR,Luo_2019_CVPR,wang_2020_ECCV}. Moreover, equipped with different pseudo label strategies, our method can improve their result and even achieve new state-of-the-art results.

\begin{figure*}
  \begin{center}
  \centering

    \includegraphics[width=0.88\linewidth]{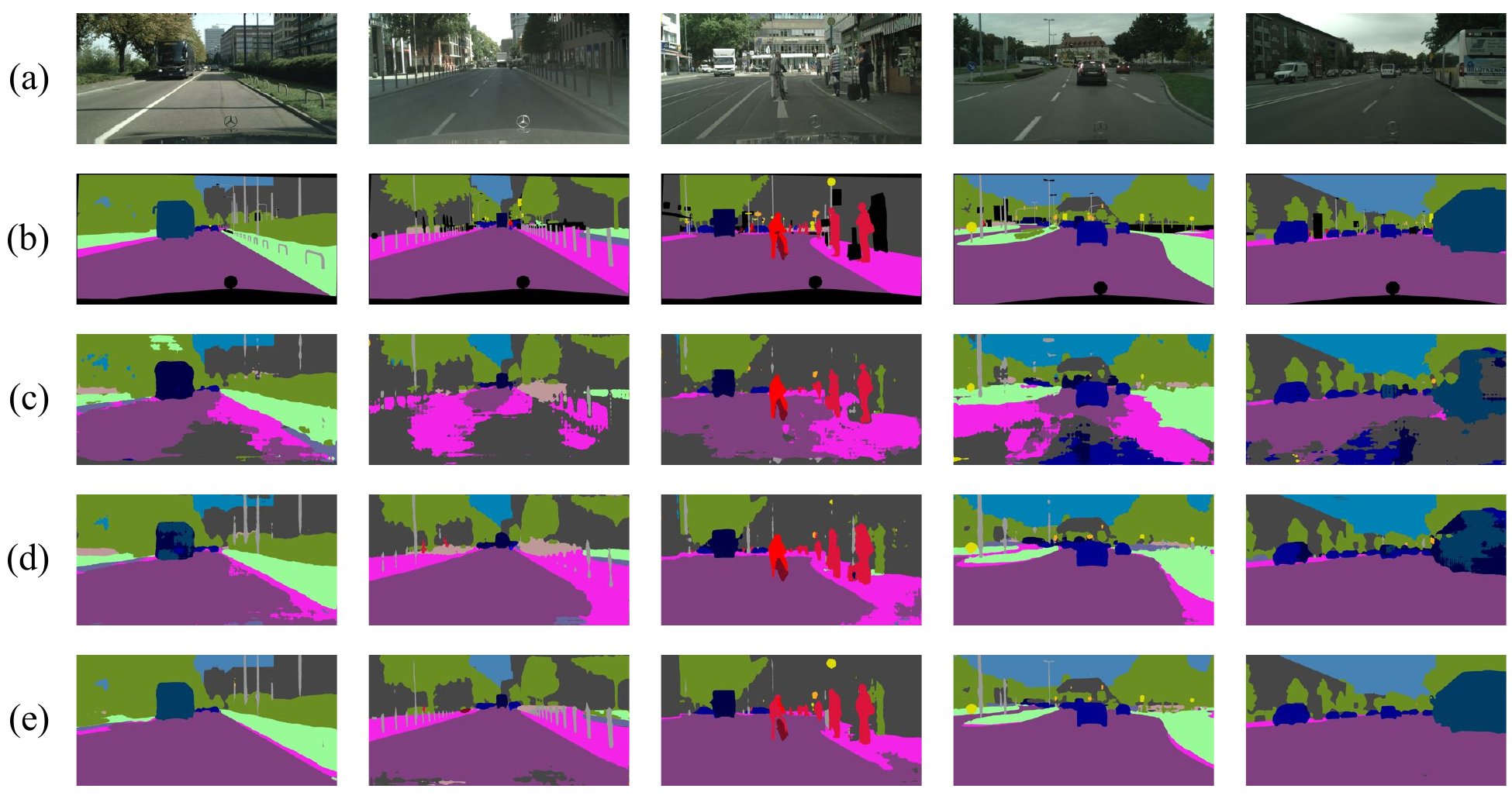}
    \caption{ Qualitative segmentation results on GTA5 to Cityscapes. We present~(a)~Target Image, (b)~Ground Truth, (c)~Source Only, (d)~Baseline~\cite{wang_2020_ECCV}, (e)~Ours.}
  \label{fig:da_seg_vis}
  \vspace{-2mm}
\end{center}
\end{figure*}

Specifically, in the scenario of GTA5 to Cityscapes, the proposed inference adjustment~(IA) and classifier refinement~(CR) achieve $48.8\%$ mIoU and $49.0\%$ mIoU, respectively. They outperform all the previous domain alignment methods by a notable margin. Our work can improve the performance for those categories with large different densities between two domains, \eg, ``truck'' and ``bike''. Combined with the self-training methods, \eg SD~\cite{wang_2020_ECCV}, IAST~\cite{mei_2020_ECCV} and ProDA~\cite{Zhang_2021_CVPR}, we achieve $52.7\%$, $55.5\%$ and $59.3\%$ mIoU, reaching new state-of-the-art results, respectively. Since the CR and IA are essentially same, we only use CR in combination with self-training methods. CLS~\cite{Liuxiao_2021_ICCV} reports the result based on IAST, and the experimental results clearly show that our work can offer a mIoU gain by $2.5\%$ with the same pseudo label method~(\ie IAST). The qualitative examples of the segmentation results of our methods are presented in Fig.~\ref{fig:da_seg_vis}, and we can observe that our method can predict a more accurate segmentation map. 
The similar results for SYNTHIA to Cityscapes can also be observed in Table~\ref{Tab:SYNTHIA}. For this task, we report the mIoU on 16 classes and 13 classes (excluding ``wall”, ``fence”, ``pole”).
Combined with the self-training methods, we can achieve $56.7\%$ and $64.5\%$ mIoU over $16$ and $13$ categories respectively, which shows that our work can outperform the existing methods on both datasets by a notable margin. This again validates the effectiveness of our method that undoes the damage of label shift for CDSS. 
 
\paragraph{Combining with self-training}
Since the self-training methods~\cite{wang_2020_ECCV,mei_2020_ECCV,Zhang_2021_CVPR} generally use domain alignment methods as their warm-up model to produce pseudo labels, our model can be easily plugged into them. We only show the result of classifier refinement in this part for conciseness. 
As shown in Table~\ref{tab:Connecting with self-training}, we first combine the self distillation~(SD)~\cite{wang_2020_ECCV} with our methods, and we can increase our baseline $49.2\%$ mIoU to $52.7\%$ mIoU. By combining our method with advanced self-training methods~\cite{mei_2020_ECCV}, we achieve a higher mIoU~($55.5\%$ \vs $53.2\%$). With ProDA~\cite{Zhang_2021_CVPR}, our model reaches $59.3\%$ mIoU, leading to a new state-of-the-art result.

\begin{table}[t]
\centering
\caption{The results~(in $\%$) of combining with different self-training methods on GTA5 to Cityscapes. }
\resizebox{0.7\linewidth}{!}{
\begin{tabular}{ccc}
\hline
  methods      & mIoU  &   $\Delta$   \\ \hline\hline
Baseline+SD~\cite{wang_2020_ECCV}       & 49.2 & \\
Ours+SD& 52.7 & 3.5$\uparrow$ \\\hline
IAST~\cite{mei_2020_ECCV} & 51.5 &   \\ 
Baseline+IAST & 53.2 &   \\
Ours+IAST & 55.5 & 2.3$\uparrow$ \\\hline
ProDA~\cite{Zhang_2021_CVPR} stage1 & 53.7 &   \\
Baseline+ProDA stage1 &  55.1 &  \\
Ours+ProDA stage1 & 57.6 &  2.5$\uparrow$ \\\hline
ProDA  & 57.5 &   \\
Ours+ProDA & 59.3 & 1.8$\uparrow$ \\\hline

\end{tabular}
}
\label{tab:Connecting with self-training} 
\vspace{-5mm}
\end{table}

\paragraph{Classifier agnostic} The baseline method FADA~\cite{wang_2020_ECCV} builds upon the original ASPP, and we reproduce the modified ASPP version for comparisons. As shown in Table~\ref{tab:Classifier Agnostic}, the proposed method can improve the baseline by $1.1\% \sim 2.1\%$ mIoU in two different classifiers. Considering various semantic segmentation classifiers, we believe our methods can be applied to any other type of classifier.

\begin{table}[t]
\centering
\caption{The results~(in $\%$) with different classifiers on GTA5 to Cityscapes. }
\resizebox{0.7\linewidth}{!}{
\begin{tabular}{cccc}
\hline
  methods      & classifier & mIoU  &   $\Delta$   \\ \hline\hline
FADA~\cite{wang_2020_ECCV} & Original ASPP      & 46.9 & \\
Our IA& Original ASPP & 48.8 & 1.9$\uparrow$ \\
Our CR& Original ASPP & 49.0 & 2.1$\uparrow$ \\ \hline
FADA~\cite{wang_2020_ECCV} & Modified ASPP      & 47.6 & \\
Our IA& Modified ASPP & 49.2 & 1.6$\uparrow$ \\
Our CR& Modified ASPP & 48.7 & 1.1$\uparrow$ \\\hline
\vspace{-10mm}
\end{tabular}
}
\label{tab:Classifier Agnostic} 
\end{table}

\section{Conclusion}
In this paper, we tackle the label shift problem for the CDSS task. We show in depth that label shift often causes the classifier bias problem in a learnt model, which, however, can be effectively avoided by aligning the data conditional distribution and correcting the posterior probability. For that, we employ class-level domain alignment for aligning conditional distribution and propose to correct the classifier bias from the source domain to the target domain by remolding the classifier predictions. As demonstrated in the experiments, our proposed approach achieves new state-of-the-art performance and outperforms all existing methods by a notable margin on two CDSS benchmark settings, showing the importance of undoing the damage of label shift for CDSS.

\noindent
\textbf{Limitation:} This work mainly focus on the close-set cross-domain semantic segmentation task. The open-set domain adaptation and partial domain adaptation setting should be considered in the future. 

\noindent
\textbf{Acknowledgement:} This work is supported by the Major Project for New Generation of AI under Grant No. 2018AAA0100400, the National Natural Science Foundation of China~(Grant No. 62176047) and Beijing Natural Science Foundation~(Z190023).


{\small
\bibliographystyle{ieee_fullname}
\bibliography{egbib}

\begin{thebibliography}{10}\itemsep=-1pt

\bibitem{badrinarayanan2017segnet}
Vijay Badrinarayanan, Alex Kendall, and Roberto Cipolla.
\newblock Segnet: A deep convolutional encoder-decoder architecture for image
  segmentation.
\newblock {\em IEEE transactions on pattern analysis and machine intelligence},
  39(12):2481--2495, 2017.

\bibitem{ben2010theory}
Shai Ben-David, John Blitzer, Koby Crammer, Alex Kulesza, Fernando Pereira, and
  Jennifer~Wortman Vaughan.
\newblock A theory of learning from different domains.
\newblock {\em Machine learning}, 79(1):151--175, 2010.

\bibitem{ben2007analysis}
Shai Ben-David, John Blitzer, Koby Crammer, Fernando Pereira, et~al.
\newblock Analysis of representations for domain adaptation.
\newblock {\em Advances in neural information processing systems}, 19:137,
  2007.

\bibitem{byrd2019effect}
Jonathon Byrd and Zachary Lipton.
\newblock What is the effect of importance weighting in deep learning?
\newblock In {\em International Conference on Machine Learning}, pages
  872--881. PMLR, 2019.

\bibitem{Caesar_2018_CVPR}
Holger Caesar, Jasper Uijlings, and Vittorio Ferrari.
\newblock Coco-stuff: Thing and stuff classes in context.
\newblock In {\em Proceedings of the IEEE Conference on Computer Vision and
  Pattern Recognition (CVPR)}, June 2018.

\bibitem{10.5555/1642293.1642455}
Yee~Seng Chan and Hwee~Tou Ng.
\newblock Word sense disambiguation with distribution estimation.
\newblock In {\em Proceedings of the 19th International Joint Conference on
  Artificial Intelligence}, IJCAI'05, page 1010–1015, San Francisco, CA, USA,
  2005. Morgan Kaufmann Publishers Inc.

\bibitem{chen2019progressive}
Chaoqi Chen, Weiping Xie, Wenbing Huang, Yu Rong, Xinghao Ding, Yue Huang,
  Tingyang Xu, and Junzhou Huang.
\newblock Progressive feature alignment for unsupervised domain adaptation.
\newblock In {\em Proceedings of the IEEE/CVF Conference on Computer Vision and
  Pattern Recognition}, pages 627--636, 2019.

\bibitem{chen2017deeplab}
Liang-Chieh Chen, George Papandreou, Iasonas Kokkinos, Kevin Murphy, and
  Alan~L. Yuille.
\newblock Deeplab: Semantic image segmentation with deep convolutional nets,
  atrous convolution, and fully connected crfs.
\newblock {\em IEEE Transactions on Pattern Analysis and Machine Intelligence},
  40(4):834--848, 2018.

\bibitem{chen2017rethinking}
Liang-Chieh Chen, George Papandreou, Florian Schroff, and Hartwig Adam.
\newblock Rethinking atrous convolution for semantic image segmentation.
\newblock {\em arXiv preprint arXiv:1706.05587}, 2017.

\bibitem{chen2018domain}
Yuhua Chen, Wen Li, Christos Sakaridis, Dengxin Dai, and Luc Van~Gool.
\newblock Domain adaptive faster r-cnn for object detection in the wild.
\newblock In {\em CVPR}, 2018.

\bibitem{Chen_2018_CVPR}
Yuhua Chen, Wen Li, and Luc Van~Gool.
\newblock Road: Reality oriented adaptation for semantic segmentation of urban
  scenes.
\newblock In {\em The IEEE Conference on Computer Vision and Pattern
  Recognition (CVPR)}, June 2018.

\bibitem{DMCD}
Tong Chu, Yahao Liu, Jinhong Deng, Wen Li, and Lixin Duan.
\newblock Denoised maximum classifier discrepancy for source-free unsupervised
  domain adaptation.
\newblock In {\em Thirty-Sixth AAAI Conference on Artificial Intelligence
  (AAAI-22)}, 2022.

\bibitem{cordts2016cityscapes}
Marius Cordts, Mohamed Omran, Sebastian Ramos, Timo Rehfeld, Markus Enzweiler,
  Rodrigo Benenson, Uwe Franke, Stefan Roth, and Bernt Schiele.
\newblock The cityscapes dataset for semantic urban scene understanding.
\newblock In {\em Proceedings of the IEEE Conference on Computer Vision and
  Pattern Recognition (CVPR)}, June 2016.

\bibitem{deng2009imagenet}
Jia Deng, Wei Dong, Richard Socher, Li-Jia Li, Kai Li, and Li Fei-Fei.
\newblock Imagenet: A large-scale hierarchical image database.
\newblock In {\em 2009 IEEE Conference on Computer Vision and Pattern
  Recognition}, pages 248--255, 2009.

\bibitem{Deng_2021_CVPR}
Jinhong Deng, Wen Li, Yuhua Chen, and Lixin Duan.
\newblock Unbiased mean teacher for cross-domain object detection.
\newblock In {\em Proceedings of the IEEE/CVF Conference on Computer Vision and
  Pattern Recognition (CVPR)}, pages 4091--4101, June 2021.

\bibitem{9616392_Dong}
Jiahua Dong, Yang Cong, Gan Sun, Zhen Fang, and Zhengming Ding.
\newblock Where and how to transfer: Knowledge aggregation-induced
  transferability perception for unsupervised domain adaptation.
\newblock {\em IEEE Transactions on Pattern Analysis and Machine Intelligence},
  pages 1--1, 2021.

\bibitem{What_Transferred_Dong_CVPR2020}
Jiahua Dong, Yang Cong, Gan Sun, Bineng Zhong, and Xiaowei Xu.
\newblock What can be transferred: Unsupervised domain adaptation for
  endoscopic lesions segmentation.
\newblock In {\em IEEE/CVF Conference on Computer Vision and Pattern
  Recognition (CVPR)}, pages 4022--4031, June 2020.

\bibitem{Du_2019_ICCV}
Liang Du, Jingang Tan, Hongye Yang, Jianfeng Feng, Xiangyang Xue, Qibao Zheng,
  Xiaoqing Ye, and Xiaolin Zhang.
\newblock {SSF-DAN}: Separated semantic feature based domain adaptation network
  for semantic segmentation.
\newblock In {\em Proceedings of the IEEE/CVF International Conference on
  Computer Vision (ICCV)}, October 2019.

\bibitem{10.1016/j.neunet.2013.11.010}
Marthinus~Christoffel Du~Plessis and Masashi Sugiyama.
\newblock Semi-supervised learning of class balance under class-prior change by
  distribution matching.
\newblock {\em Neural Netw.}, 50:110–119, Feb. 2014.

\bibitem{ganin2016domain}
Yaroslav Ganin, Evgeniya Ustinova, Hana Ajakan, Pascal Germain, Hugo
  Larochelle, Fran{\c{c}}ois Laviolette, Mario Marchand, and Victor Lempitsky.
\newblock Domain-adversarial training of neural networks.
\newblock {\em The journal of machine learning research}, 17(1):2096--2030,
  2016.

\bibitem{guo2021metacorrection}
Xiaoqing Guo, Chen Yang, Baopu Li, and Yixuan Yuan.
\newblock Metacorrection: Domain-aware meta loss correction for unsupervised
  domain adaptation in semantic segmentation.
\newblock In {\em IEEE Conference on Computer Vision and Pattern Recognition
  (CVPR)}, 2021.

\bibitem{he2016deep}
Kaiming He, Xiangyu Zhang, Shaoqing Ren, and Jian Sun.
\newblock Deep residual learning for image recognition.
\newblock In {\em Proceedings of the IEEE Conference on Computer Vision and
  Pattern Recognition (CVPR)}, June 2016.

\bibitem{hoffman2016fcns}
Judy Hoffman, Dequan Wang, Fisher Yu, and Trevor Darrell.
\newblock Fcns in the wild: Pixel-level adversarial and constraint-based
  adaptation.
\newblock {\em arXiv preprint arXiv:1612.02649}, 2016.

\bibitem{kumar2018co}
Abhishek Kumar, Prasanna Sattigeri, Kahini Wadhawan, Leonid Karlinsky, Rogerio
  Feris, William~T Freeman, and Gregory Wornell.
\newblock Co-regularized alignment for unsupervised domain adaptation.
\newblock {\em arXiv preprint arXiv:1811.05443}, 2018.

\bibitem{Liuxiao_2021_ICCV}
Xiaofeng Liu, Zhenhua Guo, Site Li, Fangxu Xing, Jane You, C.-C.~Jay Kuo,
  Georges El~Fakhri, and Jonghye Woo.
\newblock Adversarial unsupervised domain adaptation with conditional and label
  shift: Infer, align and iterate.
\newblock In {\em Proceedings of the IEEE/CVF International Conference on
  Computer Vision (ICCV)}, pages 10367--10376, October 2021.

\bibitem{Liu_2021_ICCV}
Yahao Liu, Jinhong Deng, Xinchen Gao, Wen Li, and Lixin Duan.
\newblock {BAPA-Net}: Boundary adaptation and prototype alignment for
  cross-domain semantic segmentation.
\newblock In {\em Proceedings of the IEEE/CVF International Conference on
  Computer Vision (ICCV)}, pages 8801--8811, October 2021.

\bibitem{long2015fully}
Jonathan Long, Evan Shelhamer, and Trevor Darrell.
\newblock Fully convolutional networks for semantic segmentation.
\newblock In {\em Proceedings of the IEEE Conference on Computer Vision and
  Pattern Recognition (CVPR)}, June 2015.

\bibitem{long2017conditional}
Mingsheng Long, ZHANGJIE CAO, Jianmin Wang, and Michael~I Jordan.
\newblock Conditional adversarial domain adaptation.
\newblock In S. Bengio, H. Wallach, H. Larochelle, K. Grauman, N. Cesa-Bianchi,
  and R. Garnett, editors, {\em Advances in Neural Information Processing
  Systems}, volume~31. Curran Associates, Inc., 2018.

\bibitem{Luo_2019_CVPR}
Yawei Luo, Liang Zheng, Tao Guan, Junqing Yu, and Yi Yang.
\newblock Taking a closer look at domain shift: Category-level adversaries for
  semantics consistent domain adaptation.
\newblock In {\em The IEEE Conference on Computer Vision and Pattern
  Recognition (CVPR)}, June 2019.

\bibitem{luo2021conditional}
You-Wei Luo and Chuan-Xian Ren.
\newblock Conditional bures metric for domain adaptation.
\newblock In {\em Proceedings of the IEEE/CVF Conference on Computer Vision and
  Pattern Recognition}, pages 13989--13998, 2021.

\bibitem{Lv_2020_CVPR}
Fengmao Lv, Tao Liang, Xiang Chen, and Guosheng Lin.
\newblock Cross-domain semantic segmentation via domain-invariant interactive
  relation transfer.
\newblock In {\em Proceedings of the IEEE/CVF Conference on Computer Vision and
  Pattern Recognition (CVPR)}, June 2020.

\bibitem{mei_2020_ECCV}
Ke Mei, Chuang Zhu, Jiaqi Zou, and Shanghang Zhang.
\newblock Instance adaptive self-training for unsupervised domain adaptation.
\newblock In {\em The European Conference on Computer Vision (ECCV)}, 2020.

\bibitem{menon2021longtail}
Aditya~Krishna Menon, Sadeep Jayasumana, Ankit~Singh Rawat, Himanshu Jain,
  Andreas Veit, and Sanjiv Kumar.
\newblock Long-tail learning via logit adjustment.
\newblock In {\em International Conference on Learning Representations}, 2021.

\bibitem{Paul_WeakSegDA_ECCV20}
Sujoy Paul, Yi-Hsuan Tsai, Samuel Schulter, Amit~K. Roy-Chowdhury, and Manmohan
  Chandraker.
\newblock Domain adaptive semantic segmentation using weak labels.
\newblock In {\em European Conference on Computer Vision (ECCV)}, 2020.

\bibitem{pinheiro2015image}
Pedro~O Pinheiro and Ronan Collobert.
\newblock From image-level to pixel-level labeling with convolutional networks.
\newblock In {\em Proceedings of the IEEE conference on computer vision and
  pattern recognition}, pages 1713--1721, 2015.

\bibitem{richter2016playing}
Stephan~R. Richter, Vibhav Vineet, Stefan Roth, and Vladlen Koltun.
\newblock Playing for data: {G}round truth from computer games.
\newblock In Bastian Leibe, Jiri Matas, Nicu Sebe, and Max Welling, editors,
  {\em European Conference on Computer Vision (ECCV)}, volume 9906 of {\em
  LNCS}, pages 102--118. Springer International Publishing, 2016.

\bibitem{ronneberger2015u}
Olaf Ronneberger, Philipp Fischer, and Thomas Brox.
\newblock U-net: Convolutional networks for biomedical image segmentation.
\newblock In {\em International Conference on Medical image computing and
  computer-assisted intervention}, pages 234--241. Springer, 2015.

\bibitem{ros2016synthia}
German Ros, Laura Sellart, Joanna Materzynska, David Vazquez, and Antonio~M.
  Lopez.
\newblock The synthia dataset: A large collection of synthetic images for
  semantic segmentation of urban scenes.
\newblock In {\em Proceedings of the IEEE Conference on Computer Vision and
  Pattern Recognition (CVPR)}, June 2016.

\bibitem{saito2018maximum}
Kuniaki Saito, Kohei Watanabe, Yoshitaka Ushiku, and Tatsuya Harada.
\newblock Maximum classifier discrepancy for unsupervised domain adaptation.
\newblock In {\em Proceedings of the IEEE conference on computer vision and
  pattern recognition}, pages 3723--3732, 2018.

\bibitem{tachet2020domain}
Remi Tachet~des Combes, Han Zhao, Yu-Xiang Wang, and Geoffrey~J Gordon.
\newblock Domain adaptation with conditional distribution matching and
  generalized label shift.
\newblock {\em Advances in Neural Information Processing Systems}, 33, 2020.

\bibitem{tranheden2021dacs}
Wilhelm Tranheden, Viktor Olsson, Juliano Pinto, and Lennart Svensson.
\newblock {DACS}: Domain adaptation via cross-domain mixed sampling.
\newblock In {\em Proceedings of the IEEE/CVF Winter Conference on Applications
  of Computer Vision (WACV)}, pages 1379--1389, January 2021.

\bibitem{Tsai_2018_CVPR}
Yi-Hsuan Tsai, Wei-Chih Hung, Samuel Schulter, Kihyuk Sohn, Ming-Hsuan Yang,
  and Manmohan Chandraker.
\newblock Learning to adapt structured output space for semantic segmentation.
\newblock In {\em The IEEE Conference on Computer Vision and Pattern
  Recognition (CVPR)}, June 2018.

\bibitem{Vu_2019_CVPR}
Tuan-Hung Vu, Himalaya Jain, Maxime Bucher, Matthieu Cord, and Patrick Perez.
\newblock {ADVENT}: Adversarial entropy minimization for domain adaptation in
  semantic segmentation.
\newblock In {\em The IEEE Conference on Computer Vision and Pattern
  Recognition (CVPR)}, June 2019.

\bibitem{wang_2020_ECCV}
Haoran Wang, Tong Shen, Wei Zhang, Lingyu Duan, and Tao Mei.
\newblock Classes matter: A fine-grained adversarial approach to cross-domain
  semantic segmentation.
\newblock In {\em The European Conference on Computer Vision (ECCV)}, 2020.

\bibitem{wang2019deep}
Jingdong Wang, Ke Sun, Tianheng Cheng, Borui Jiang, Chaorui Deng, Yang Zhao,
  Dong Liu, Yadong Mu, Mingkui Tan, Xinggang Wang, Wenyu Liu, and Bin Xiao.
\newblock Deep high-resolution representation learning for visual recognition.
\newblock {\em TPAMI}, 2019.

\bibitem{xie2021segformer}
Enze Xie, Wenhai Wang, Zhiding Yu, Anima Anandkumar, Jose~M Alvarez, and Ping
  Luo.
\newblock Segformer: Simple and efficient design for semantic segmentation with
  transformers.
\newblock {\em arXiv preprint arXiv:2105.15203}, 2021.

\bibitem{Yang_2020_CVPR}
Yanchao Yang and Stefano Soatto.
\newblock {FDA}: Fourier domain adaptation for semantic segmentation.
\newblock In {\em Proceedings of the IEEE/CVF Conference on Computer Vision and
  Pattern Recognition (CVPR)}, June 2020.

\bibitem{YuanCW19}
Yuhui Yuan, Xilin Chen, and Jingdong Wang.
\newblock Object-contextual representations for semantic segmentation.
\newblock In {\em ECCV}, 2020.

\bibitem{Zhang_2021_CVPR}
Pan Zhang, Bo Zhang, Ting Zhang, Dong Chen, Yong Wang, and Fang Wen.
\newblock Prototypical pseudo label denoising and target structure learning for
  domain adaptive semantic segmentation.
\newblock In {\em Proceedings of the IEEE/CVF Conference on Computer Vision and
  Pattern Recognition (CVPR)}, pages 12414--12424, June 2021.

\bibitem{zhao2017pyramid}
Hengshuang Zhao, Jianping Shi, Xiaojuan Qi, Xiaogang Wang, and Jiaya Jia.
\newblock Pyramid scene parsing network.
\newblock In {\em Proceedings of the IEEE Conference on Computer Vision and
  Pattern Recognition (CVPR)}, July 2017.

\bibitem{zheng2019unsupervised}
Zhedong Zheng and Yi Yang.
\newblock Unsupervised scene adaptation with memory regularization in vivo.
\newblock In {\em IJCAI}, 2020.

\bibitem{zheng_2020_jour}
Zhedong Zheng and Yi Yang.
\newblock Rectifying pseudo label learning via uncertainty estimation for
  domain adaptive semantic segmentation.
\newblock {\em International Journal of Computer Vision}, 129(4):1106--1120,
  2021.

\bibitem{8100027}
Bolei Zhou, Hang Zhao, Xavier Puig, Sanja Fidler, Adela Barriuso, and Antonio
  Torralba.
\newblock Scene parsing through ade20k dataset.
\newblock In {\em 2017 IEEE Conference on Computer Vision and Pattern
  Recognition (CVPR)}, pages 5122--5130, 2017.

\bibitem{Zou_2018_ECCV}
Yang Zou, Zhiding Yu, BVK Kumar, and Jinsong Wang.
\newblock Unsupervised domain adaptation for semantic segmentation via
  class-balanced self-training.
\newblock In {\em Proceedings of the European conference on computer vision
  (ECCV)}, pages 289--305, 2018.

\bibitem{Zou_2019_ICCV}
Yang Zou, Zhiding Yu, Xiaofeng Liu, B.V.K.~Vijaya Kumar, and Jinsong Wang.
\newblock Confidence regularized self-training.
\newblock In {\em Proceedings of the IEEE/CVF International Conference on
  Computer Vision (ICCV)}, October 2019.

\end{thebibliography}
}

\end{document}